\documentclass[sigconf]{acmart}
\usepackage[capitalise]{cleveref}
\usepackage{subcaption}
\usepackage{multirow}
\usepackage{rotating}
\usepackage{tablefootnote}
\usepackage{ctable}
\usepackage{threeparttable}
\AtBeginDocument{%
  }

\copyrightyear{2025}
\acmYear{2025}
\setcopyright{cc}
\setcctype{by-nc}
\acmConference[ISWC '25]{Proceedings of the 2025 ACM International Symposium on Wearable Computers}{October 12--16, 2025}{Espoo, Finland}
\acmBooktitle{Proceedings of the 2025 ACM International Symposium on Wearable Computers (ISWC '25), October 12--16, 2025, Espoo, Finland}\acmDOI{10.1145/3715071.3750425}
\acmISBN{979-8-4007-1481-8/2025/10}

\begin{document}

\title{From Neck to Head: Bio-Impedance Sensing for Head Pose Estimation}

\author{Mengxi Liu}
\affiliation{%
  \institution{German Research Center for Artificial Intelligence}
  \city{Kaiserslautern}
  \country{Germany}
}
\email{mengxi.liu@dfki.de}

\author{Lala Shakti Swarup Ray}
\affiliation{%
  \institution{German Research Center for Artificial Intelligence}
  \city{Kaiserslautern}
  \country{Germany}
}
\email{lala_shakti_swarup.ray@dfki.de}

\author{Sizhen Bian}
\affiliation{%
  \institution{German Research Center for Artificial Intelligence}
  \city{Kaiserslautern}
  \country{Germany}
}
\email{sizhen.bian@dfki.de}

\author{Ko Watanabe}
\affiliation{%
  \institution{German Research Center for Artificial Intelligence}
  \city{Kaiserslautern}
  \country{Germany}
}
\email{ko.watanabe@dfki.de}

\author{Ankur Bhatt}
\affiliation{%
  \institution{German Research Center for Artificial Intelligence}
  \city{Kaiserslautern}
  \country{Germany}
}
\email{ankur.bhatt@dfki.de}

\author{Joanna Sorysz}
\affiliation{%
  \institution{German Research Center for Artificial Intelligence}
  \city{Kaiserslautern}
  \country{Germany}
}
\email{joanna.sorysz@dfki.de}

\author{Russel Torah}
\affiliation{%
  \institution{University of Southampton}
  \city{Southampton}
  \country{United Kingdom}
}
\email{rnt@ecs.soton.ac.uk}

\author{Bo Zhou}
\affiliation{%
  \institution{German Research Center for Artificial Intelligence}
  \city{Kaiserslautern}
  \country{Germany}
}
\email{bo.zhou@dfki.de}

\author{Paul Lukowicz}
\affiliation{%
  \institution{German Research Center for Artificial Intelligence}
  \city{Kaiserslautern}
  \country{Germany}
}
\email{paul.lukowicz@dfki.de}

\renewcommand{\shortauthors}{Mengxi Liu et al.}

\begin{abstract}

We present NeckSense, a novel wearable system for head pose tracking that leverages multi-channel bio-impedance sensing with soft, dry electrodes embedded in a lightweight, necklace-style form factor. NeckSense captures dynamic changes in tissue impedance around the neck, which are modulated by head rotations and subtle muscle activations. 
To robustly estimate head pose, we propose a deep learning framework that integrates anatomical priors, including joint constraints and natural head rotation ranges, into the loss function design. We validated NeckSense on 7 participants using the current state-of-the-art pose estimation model as ground truth. Our system achieves a mean per-vertex error of 25.9 mm across various head movements with a leave-one-person-out cross-validation method, demonstrating that a compact bio-impedance wearable can deliver head-tracking performance comparable to state-of-the-art vision-based methods.

\end{abstract}




\maketitle

\section{Introduction}


Continuous head pose tracking is essential for applications like AR/VR and human-robot interaction \cite{murphy2010head,liu2022arhpe,zimmermann20183d}. 
However, existing tracking technologies face challenges for real-world use. 
Vision-based systems are compromised by occlusion and privacy concerns, while common wearables present their own trade-offs. Inertial Measurement Units (IMUs) suffer from drift, and electromyography (EMG), while effective for gesture recognition systems like the Meta Reality Labs wristband \cite{meta2024semg}, relies on active muscle contractions, making it less reliable for tracking sustained or passive postures \cite{disselhorst2009surface}. This highlights a clear need for a modality that can robustly infer pose from the body's underlying anatomy, independent of line-of-sight or the nature of muscle activation.

To address this gap, we propose a novel approach using bio-impedance sensing. Our method is fundamentally distinct from and more efficient than the existing Electrical Impedance Tomography (EIT) in pose/gesture estimation study \cite{zhang2015tomo,zhang2016advancing,kyu2024eitpose}. Whereas EIT aims to reconstruct a full internal conductivity image using complex hardware (often 8+ electrodes) and computationally intensive algorithms, and the pose/gesture is extracted from large number measurements of time-domain information (current and voltage) due to the complex EIT measurement protocol \cite{dimas2024advances}, for example, there are 40 channels in EITPose \cite{kyu2024eitpose}, we argue that this is unnecessary for the head pose estimation. Instead, our system bypasses imaging entirely. We use a simpler five-electrode configuration (one common electrode generates stimuli and four measurement channels) to directly measure frequency-domain impedance features (magnitude and phase).
The measured impedance is influenced by several factors, including the composition of underlying tissues (e.g., muscle, fat, and skin) \cite{sanchez2017electrical}, the geometry and spacing of the electrodes\cite{briko2021determination}, and the orientation of muscle fibers or structural deformation \cite{stupin2021bioimpedance}.
This non-tomographic approach allows us to robustly track both dynamic and static poses with significantly reduced hardware and algorithmic complexity.


We present NeckSense, a lightweight necklace with soft, dry, reusable electrodes for comfortable, continuous wear. Paired with a transformer-based model, Imp2Head, it maps raw impedance signals to 3D head pose while incorporating anatomical constraints to ensure physically plausible predictions.
To evaluate performance, we conducted a user study (N=7) using naturalistic motion scenarios like conversation and AR/VR interaction, with ground truth from a state-of-the-art vision-based model. NeckSense achieved a mean per-vertex error of 25.9 mm, comparable to the vision system’s own error, demonstrating it as a competitive alternative.

Overall, our contributions are: 

\begin{itemize}
\item We introduce NeckSense, the first wearable system for continuous head pose estimation using multi-channel bio-impedance sensing. It integrates five soft, reusable electrodes into a lightweight, comfortable, and camera-free necklace form factor.

\item We develop a deep learning framework that maps impedance signals to 3D head pose, incorporating biomechanical constraints (e.g., joint limits and rotation ranges) to improve anatomical plausibility and robustness.

\item We validate NeckSense on 7 participants using a camera-based motion capture system as ground truth, achieving a mean per-vertex error of 25.9 mm, comparable to state-of-the-art vision-based methods without requiring line-of-sight.

\end{itemize}
\section{Related Work}

Wearable pose estimation typically employs compact sensor modalities, including IMU \cite{kim2024prediction,georgi2015recognizing,tavakoli2018robust,mollyn2023imuposer,xu2024mobileposer}, capacitive sensors \cite{ziraknejad2014vehicle,bian2024body,zhou2023mocapose}, bio-transducers, like EMG \cite{georgi2015recognizing,aguiar2017hand,tavakoli2018robust,meta2024semg,chen2021exgsense} and acoustical sensors \cite{iravantchi2019interferi}, due to their small form factor and suitability for body-mounted applications. 
IMUs estimate head orientation by integrating angular velocity and acceleration signals~\cite{lopez2016wearable}. Single-IMU setups, typically integrated into headbands or eyeglasses, are simple but susceptible to drift~\cite{Severin2020}. Multi-IMU arrays improve accuracy but add complexity, bulk, and discomfort~\cite{Breen2009}. Hybrid IMU-camera approaches partially resolve drift but are heavily influenced by environmental factors such as lighting and visibility.
EMG-based systems effectively track gestures and limb movements by capturing muscle activation signals~\cite{georgi2015recognizing,aguiar2017hand}. However, EMG inherently relies on active muscle contraction, limiting its efficacy for continuous or static head pose estimation. Recent EMG-based head tracking systems necessitate electrodes on neck and facial areas, significantly reducing comfort and social acceptability~\cite{kim2024prediction}. Additionally, variations in electrode placement can severely affect long-term reliability. 
Capacitive sensing offers another promising alternative for gesture and pose estimation \cite{bian2024body,bian2022state}, yet its susceptibility to environmental interference limits its robustness in everyday use cases~\cite{bian2024body}. Thus, existing wearable sensor solutions face inherent trade-offs among accuracy, portability, usability, and environmental robustness, motivating continued exploration of complementary sensing modalities.

Bio-impedance sensing has been extensively explored for various physiological monitoring tasks, such as respiration~\cite{liu2025ibreath}, hydration~\cite{o2002bioelectrical}, and swallowing detection~\cite{kusuhara2004impedance}. More recently, bio-impedance has emerged as a promising modality for gesture recognition in human-computer interaction~\cite{Waghmare2023,zhang2015tomo,liu2025ibreath} and activity recognition \cite{liu2024imove,liu2024ieat,liu2024iface}. Unlike EMG, bio-impedance does not require active muscle contraction, enabling continuous detection of both static and dynamic postures based on tissue geometry changes alone~\cite{sanchez2017electrical,stupin2021bioimpedance}. Furthermore, bio-impedance inherently supports multi-channel volumetric sensing, capturing spatial deformation patterns across body regions to provide richer contextual information than localized EMG or angular IMUs~\cite{zhang2015tomo}.

EIT, a specific bio-impedance approach, typically reconstructs internal conductivity distributions using boundary voltage and current measurements, requiring numerous electrodes (often more than eight) to achieve high-resolution imaging~\cite{zhang2015tomo,zhang2016advancing,kyu2024eitpose}. Traditional EIT employs a pairwise adjacent current stimulation pattern (Sheffield method), resulting in increased hardware complexity and data processing burdens due to the large number of measurements (specifically, $N(N-3)/2$ for an $N$-electrode system). Additionally, EIT’s inverse reconstruction is highly sensitive to noise, often leading to significant artifacts.


Unlike traditional EIT, NeckSense eliminates the need for full tomographic reconstruction in head pose estimation. It uses a simplified setup—one stimulation electrode and four strategically placed measurement electrodes, while directly leveraging frequency domain features (magnitude and phase) for efficient, robust pose estimation. This reduces hardware complexity, computational load, and sensitivity to noise.
By situating NeckSense within existing bio-transducer literature and explicitly clarifying its unique differences from traditional EIT, we highlight the physiological distinctiveness, practical advantages, and novelty of our simplified bio-impedance sensing approach for wearable head pose estimation and human-computer interaction.

\section{APPROACH}
\subsection{General Principle}

\begin{figure}[!htbp]
    \centering
\includegraphics[width=0.8\linewidth]{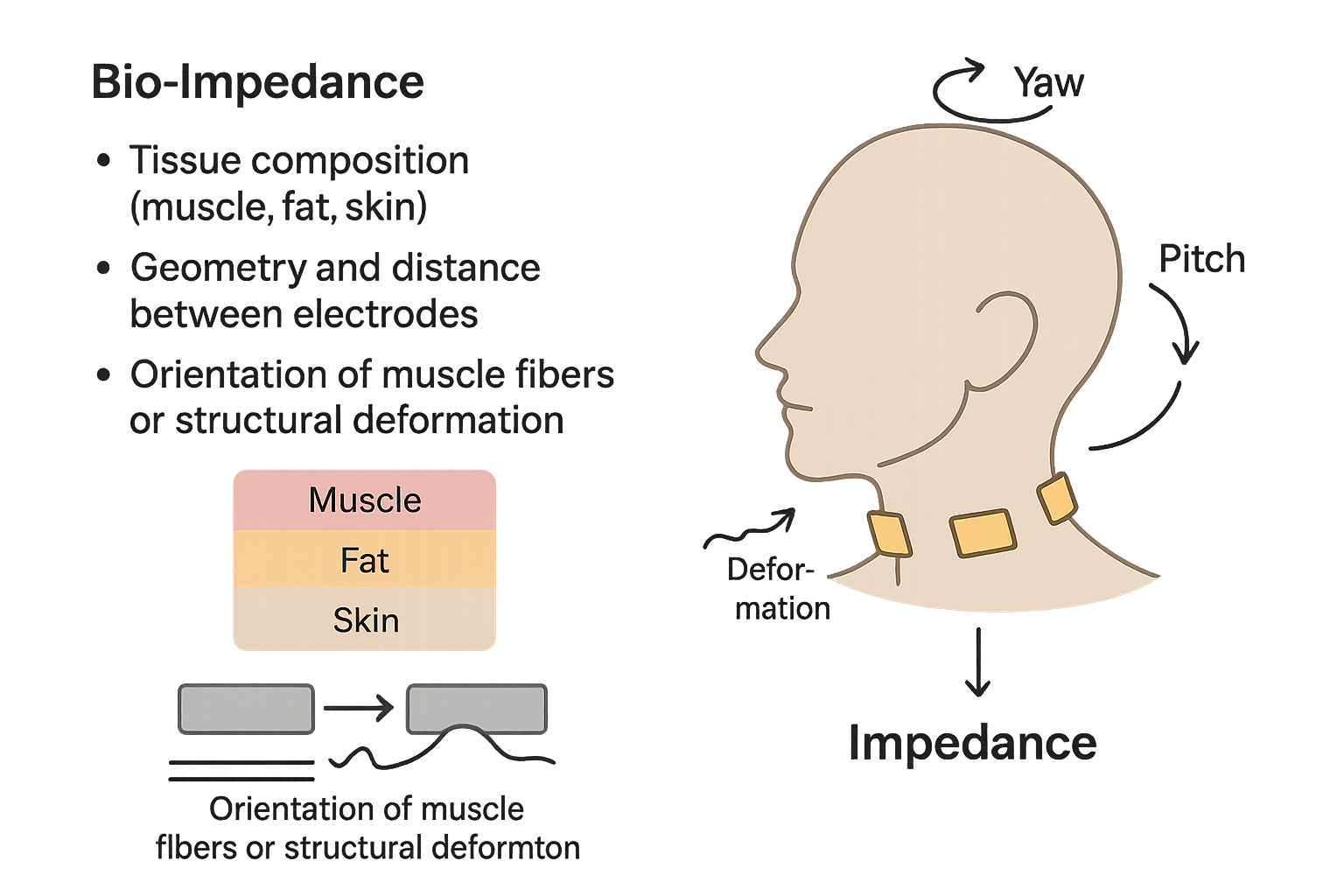}
    \caption{Bio-impedance-based head pose estimation uses five necklace-integrated electrodes to monitor neck impedance changes caused by head movement, which depend on tissue composition, electrode geometry and spacing, and muscle orientation or structural deformation}
    \label{fig:general principle}
\end{figure}

Bio-impedance sensing is an electrophysiological technique that monitors the conductive properties of biological tissue between pairs of electrodes. \cref{fig:general principle} presents the general principle of bio-impedance sensing-based head pose estimation. The measured impedance is influenced by several factors, including the composition of underlying tissues (e.g., muscle, fat, and skin) \cite{sanchez2017electrical}, the geometry and spacing of the electrodes\cite{briko2021determination}, and the orientation of muscle fibers or structural deformation \cite{stupin2021bioimpedance}.
When the head moves, particularly during yaw (turning left/right), pitch (nodding up/down), or roll (tilting sideways), several physiological changes occur. Muscular contractions and shifts in tissue can alter local conductivity, while deformation of skin and subcutaneous tissue can change the geometry and thickness of the conductive medium. Additionally, slight displacements of the skin may lead to variations in the relative positioning of the electrodes, further modulating the measured impedance.
In this work, we attach multiple electrodes around the neck and track changes in bio-impedance to estimate head pose. By leveraging the sensitivity of bio-impedance to anatomical and postural variations, our system enables camera-free, continuous head movement monitoring.

\subsection{Hardware Implementation}
\begin{figure*}[!htbp]
    \centering
    \includegraphics[width=0.7\linewidth]{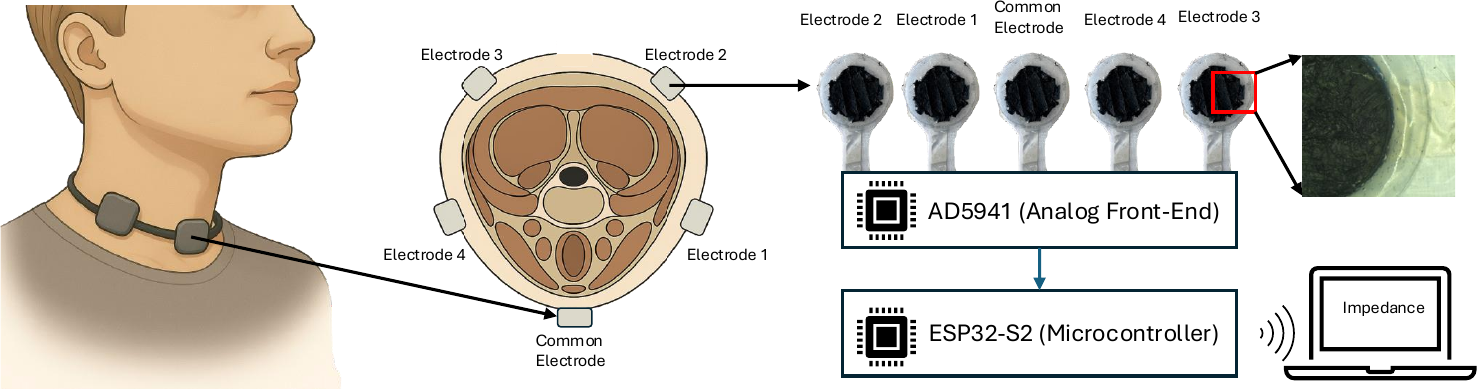}
    \caption{NeckSense uses five soft, reusable electrodes around the neck, with one for stimulation and four for impedance sensing (magnitude and phase). Signals from the AD5941 are read by an ESP32-S2 and sent via Bluetooth for real-time processing}
    \label{fig:system_design}
\end{figure*}

To efficiently acquire bio-impedance signals from the neck, we developed a compact, wearable system named NeckSense, as illustrated in \cref{fig:system_design}. The system comprises two main components: (1) a set of soft, dry electrodes integrated into a neckband, and (2) an impedance data acquisition module.

Unlike previous approaches that rely on Silver/Silver-Chloride (Ag/AgCl) wet electrodes which offer high signal quality but suffer from limitations such as gel evaporation, skin irritation, and lack of reusability, or the rigid dry electrodes lacking comfort, 
we adopt soft, dry electrodes. These are well-suited for daily-life, long-term use and can be seamlessly embedded into a flexible, wearable necklace.
The soft, dry electrodes made by screen-printed technology use conductive inks on polyester-cotton fabric for a soft, skin-conforming interface. A PET laminate layer enables clean printing of silver ink (Fabinks TC-C4007), which is cured at 120 \textdegree{C} degree and heat-pressed onto the fabric at 190 \textdegree{C}. The exposed pad is covered with a conductive silicone layer made from activated carbon black and silicone binder (Fabinks TC-E0002), stencil-printed and cured at 80°C. A second non-conductive silicone ring improves adhesion and durability. Electrodes show low resistance (<0.7 $\Omega$ silver, ~1k$\Omega$ silicone) and endure 400,000 bending cycles and 50 wash cycles, confirming long-term performance \cite{komolafe2024improving}.

The impedance acquisition module consists of two integrated circuits: an Analog front-end (AFE) chip and Microcontroller unit (MCU).
AD5941 generates the sinusoidal excitation signal, measures the resulting current, and includes an on-chip Fast Fourier Transform (FFT) accelerator to efficiently compute impedance in the frequency domain.
ESP32-S2 interfaces with the AD5941 via SPI, reads the real and imaginary impedance components, converts them to magnitude and phase, and transmits the data via Bluetooth to a host device (e.g., a computer) for further processing.
This system design allows continuous, high-resolution tracking of neck bio-impedance in real time, with minimal form factor and power consumption, making it ideal for wearable, daily use.

Since our system does not require image reconstruction, it does not strictly follow the traditional EIT measurement protocol, allowing for flexible electrode configurations. In this study, we selected a five-electrode setup based on considerations of hardware capability, pose estimation performance, and user experience as shown in \cref{fig:system_design}. The AD5941 chip supports four measurement channels without the need for an external multiplexer. Placing four electrodes at different positions around the neck allows us to effectively capture major head movement directions. While adding more electrodes could potentially enhance performance, it may also degrade user experience and increase system complexity.

\subsection{Imp2Head: Head Pose Tracking Algorithm}
\begin{figure*}[!htbp]
    \centering
    \includegraphics[width=0.7\linewidth]{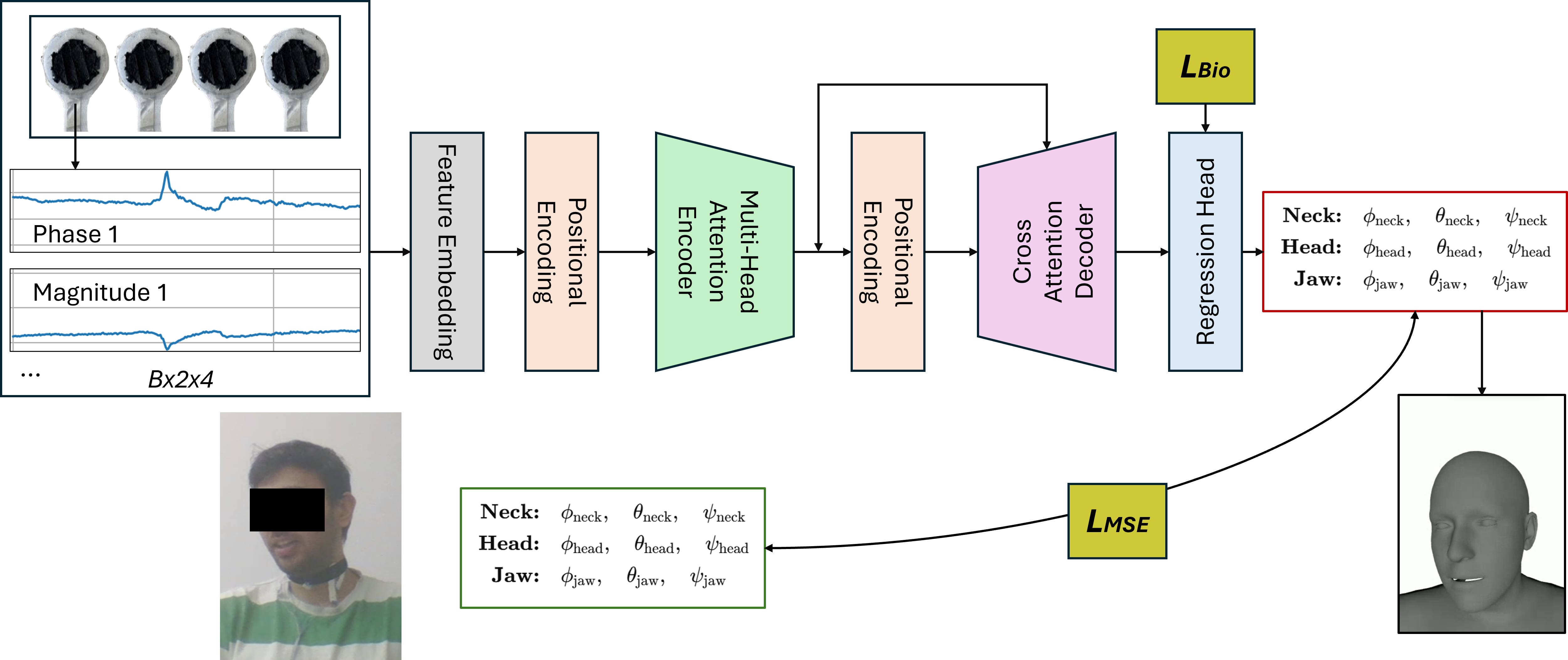}
    \caption{Architecture of the Imp2Head model, a modular encoder-decoder transformer that maps temporal impedance signals to joint rotations. The encoder captures temporal patterns, while the decoder predicts joint angles autoregressively using attention over encoded features. The loss combines mean squared error with biomechanical constraints to ensure anatomically valid predictions.}
    \label{fig:model}
\end{figure*}
This work proposes a Transformer-based sequence-to-sequence model to estimate the head pose dynamics of a person using impedance signals. The model predicts the temporal evolution of three anatomical joints: neck, head, and jaw each described using SMPL-compatible axis-angle rotations.

Ground truth joint angles were obtained by processing RGB video using the OS-X SOTA full body pose estimation pipeline \cite{lin2023one}, which extracts full-body SMPL-X parameters for each frame. 
From this output, we retained only the axis-angle rotations for the neck, head, and jaw joints. All other joints were discarded.
To reduce high-frequency jitter and ensure temporal consistency in the SMPL-X joint rotations, we apply a smoothing algorithm to the ground truth trajectories before training. Since axis-angle representations can suffer from discontinuities, we first convert each joint's rotation from axis-angle to quaternion format. 
Let $q_t \in \mathbb{R}^4$ denote the unit quaternion representing the rotation at time $t$. We apply Gaussian smoothing over a temporal window using the following equation:

\begin{equation}
    \hat{q}_t = \frac{\sum\limits_{k=-K}^{K} w_k \cdot q_{t+k}}{\left\|\sum\limits_{k=-K}^{K} w_k \cdot q_{t+k}\right\|}, \quad w_k = \exp\left(-\frac{k^2}{2\sigma^2}\right)
\end{equation}

where $q_{t+k}$ is the quaternion at frame $t+k$, $w_k$ are Gaussian weights with bandwidth parameter $\sigma$, $K$ defines the size of the temporal window, The result is normalized to ensure $\hat{q}_t$ remains a valid unit quaternion.
After smoothing, the quaternions $\hat{q}_t$ are converted back to axis-angle format to serve as the final smoothed ground truth joint rotations. 
This method preserves global rotational structure while effectively removing high-frequency noise from the estimated motion.
The model receives as input a sequence of impedance features over a fixed-length window. Specifically, for each time step, impedance measurements are recorded from 4 sensing electrodes, each providing 2 channels (magnitude and phase), yielding an 8 dimensional feature vector, and the input sequence is defined as:
\begin{equation}
    X \in \mathbb{R}^{B \times L_{\text{in}} \times 8}, \quad L_{\text{in}} = 9 \cdot L_{\text{out}}
\end{equation}

where \( B \) is the batch size and \( L_{\text{out}} \) is the desired output sequence length. The output is a predicted sequence of SMPL joint rotations:
\begin{equation}
    \hat{Y} \in \mathbb{R}^{B \times L_{\text{out}} \times 9}
\end{equation}

Each output frame contains 9 continuous values representing the pitch, yaw, and roll angles of the neck, head, and jaw joints.

\cref{fig:model} shows the architecture of the proposed Imp2Head, which follows a modular encoder-decoder transformer framework, designed to learn temporal mappings between physiological impedance signals and anatomical joint rotations.
The architecture begins with a temporal embedding of the input impedance sequence, which transforms raw sensor readings into a learned latent representation. This is achieved by applying a trainable projection layer followed by positional encoding, enabling the model to recognize temporal ordering within the input window.
The intuition behind this step is that understanding the sequence and timing of sensor signals is crucial to accurately capture dynamic head and neck movements over time.

The encoded sequence is processed by a stack of self-attention layers in the transformer encoder. These layers capture complex temporal dependencies and contextual interactions across the full input history, allowing the model to form a global understanding of the observed impedance dynamics.
This design leverages self-attention to model long-range temporal relationships, reflecting how head and neck movements involve coordinated muscle activations spread over time rather than isolated instantaneous signals.

The decoder operates autoregressively over a learned output query, independent of any ground truth input during inference. It performs cross-attention over the encoder's latent representation, allowing the decoder to condition future pose predictions on the historical context extracted from the impedance sequence.
This enables the model to forecast continuous sequences of joint rotations by conditioning each predicted step on the learned temporal context, mirroring how past movements influence future poses in a time-dependent manner.

Finally, the decoder output is projected back into the joint angle space through a regression head, producing a continuous sequence of 3D joint rotations. This architecture allows for flexible modeling of high-dimensional, time-dependent motion data and supports multi-step forecasting with attention-guided temporal reasoning.

Imp2Head is trained using a loss function that combines mean squared error (MSE) with biomechanical constraints to ensure physically plausible outputs. Let \( \theta_{\min}^{(i)} \) and \( \theta_{\max}^{(i)} \) represent the anatomical limits for joint \( i \in \{1, \dots, 9\} \). A penalty term is used to discourage violations of these limits:

\begin{equation}
    \mathcal{L}_{\text{bio}} = \frac{1}{B \cdot L_{\text{out}} \cdot 9} \sum_{b,t,i} \left[ \max(0, \theta_{\min}^{(i)} - \hat{Y}_{b,t,i})^2 + \max(0, \hat{Y}_{b,t,i} - \theta_{\max}^{(i)})^2 \right]
\end{equation}

The total loss is given by:
\begin{equation}
    \mathcal{L}_{\text{total}} = \mathcal{L}_{\text{MSE}} + \lambda \cdot \mathcal{L}_{\text{bio}}
\end{equation}

where \( \lambda \) is a tunable hyperparameter controlling the weight of the biomechanical constraint. It ensures the model's predictions are both temporally coherent and anatomically valid.

\section{Evaluation}
\subsection{Study Design}

\begin{table*}[h]
\footnotesize
\centering
\caption{Measured joint rotation ranges (radians) for 7 participants during natural head and neck movements wearing the NeckSense device.(The joint rotation ranges are computed based on the ground truth.)}
\begin{tabular}{lccc}
\hline
\textbf{Person} & \textbf{Neck (Pitch, Yaw, Roll)} & \textbf{Head (Pitch, Yaw, Roll)} & \textbf{Jaw (Pitch, Yaw, Roll)} \\
\hline
\textbf{1} & 
\([ -0.44,\, 0.50 ]\), \([ -0.92,\, 0.89 ]\), \([ -0.63,\, 0.68 ]\) & 
\([ -0.39,\, 0.47 ]\), \([ -0.51,\, 0.49 ]\), \([ -0.42,\, 0.48 ]\) & 
\([ 0.03,\, 0.48 ]\), \([ -0.13,\, 0.14 ]\), \([ -0.12,\, 0.14 ]\) \\
\textbf{2} & 
\([ -0.68,\, 0.91 ]\), \([ -0.52,\, 0.79 ]\), \([ -0.42,\, 0.59 ]\) & 
\([ -0.47,\, 0.34 ]\), \([ -0.41,\, 0.45 ]\), \([ -0.28,\, 0.40 ]\) & 
\([ 0.11,\, 0.46 ]\), \([ -0.09,\, 0.12 ]\), \([ -0.07,\, 0.08 ]\) \\
\textbf{3} & 
\([ -0.97,\, 0.47 ]\), \([ -0.77,\, 0.35 ]\), \([ -0.62,\, 0.51 ]\) & 
\([ -0.25,\, 0.51 ]\), \([ -0.47,\, 0.39 ]\), \([ -0.40,\, 0.23 ]\) & 
\([ 0.06,\, 0.48 ]\), \([ -0.15,\, 0.04 ]\), \([ -0.10,\, 0.09 ]\) \\
\textbf{4} & 
\([ -0.88,\, 0.93 ]\), \([ -0.89,\, 0.98 ]\), \([ -0.24,\, 0.51 ]\) & 
\([ -0.35,\, 0.28 ]\), \([ -0.43,\, 0.45 ]\), \([ -0.49,\, 0.46 ]\) & 
\([ 0.00,\, 0.50 ]\), \([ -0.05,\, 0.11 ]\), \([ -0.15,\, 0.12 ]\) \\
\textbf{5} & 
\([ -0.50,\, 0.39 ]\), \([ -1.03,\, 0.94 ]\), \([ -0.65,\, 0.70 ]\) & 
\([ -0.46,\, 0.48 ]\), \([ -0.33,\, 0.50 ]\), \([ -0.11,\, 0.31 ]\) & 
\([ 0.24,\, 0.51 ]\), \([ -0.16,\, -0.01 ]\), \([ -0.06,\, 0.12 ]\) \\
\textbf{6} & 
\([ -0.86,\, 0.82 ]\), \([ -0.62,\, 0.22 ]\), \([ -0.29,\, 0.41 ]\) & 
\([ -0.29,\, 0.43 ]\), \([ -0.48,\, 0.27 ]\), \([ -0.35,\, 0.37 ]\) & 
\([ 0.13,\, 0.46 ]\), \([ -0.04,\, 0.12 ]\), \([ -0.07,\, 0.06 ]\) \\
\textbf{7} & 
\([ -1.04,\, 1.01 ]\), \([ -0.38,\, 0.56 ]\), \([ -0.59,\, 0.65 ]\) & 
\([ -0.46,\, 0.19 ]\), \([ -0.51,\, 0.28 ]\), \([ -0.49,\, 0.09 ]\) & 
\([ 0.01,\, 0.45 ]\), \([ -0.06,\, 0.11 ]\), \([ -0.15,\, 0.11 ]\) \\
\hline
\end{tabular}
\label{tab:data_stats}
\end{table*}

To comprehensively evaluate the performance of the proposed NeckSense system, we conducted a user study involving 7 participants aged 21 to 31 years ($26.45 \pm 2.87$, 2 females and 5 males). Each participant was seated comfortably at a table and wore the NeckSense wearable device.
Participants were instructed to move their heads freely, performing natural yaw, pitch, and roll movements, and also performing specific movements, including looking down, up, left, right, front-left, and front-right, they were allowed to speak, drink, and use gadgets during the session, to simulate realistic use cases such as conversation or AR/VR interaction. 
A camera was positioned in front of each participant to continuously record their movements. The resulting video footage was later used to extract the ground-truth head pose for quantitative evaluation.
Each participant took part in the experiment for approximately 30 minutes, ensuring sufficient data to analyze a range of head motions and impedance responses under unconstrained conditions.
Detailed angle range of each joint per participant can be found on \cref{tab:data_stats}.

\subsection{Experimental Details}
Imp2Head is implemented using the PyTorch on a windows system with an NVIDIA RTX 4090 GPU. The model is optimized using the Adam optimizer with an initial learning rate of \(1 \times 10^{-3}\), and a batch size of 256. A step-based learning rate scheduler is employed, reducing the learning rate by a factor of 0.5 every 25 epochs to aid convergence.
We include a biomechanical constraint term in the loss function, weighted by a factor \( \lambda = 0.1 \). The following angle limits (in radians) are imposed for each joint axis (Pitch, Yaw, Roll):

\begin{itemize}
  \item Neck: \([-1.05, 1.05]\), \([-1.05, 1.05]\), \([-0.70, 0.70]\)
  \item Head: \([-0.52, 0.52]\), \([-0.79, 0.79]\), \([-0.52, 0.52]\)
  \item Jaw: \([0.0, 0.52]\), \([-0.17, 0.17]\), \([-0.17, 0.17]\)
\end{itemize}

To ensure generalization across subjects, we adopt a Leave-One-Person-Out evaluation protocol. 
This setup ensures that the model is evaluated on completely unseen individuals, which tests its ability to generalize to new users and anatomy. A sliding window approach is used to generate input-output pairs, with 90 frame input windows and 10 frame output sequences.

\subsection{Evaluation Metric}
To evaluate the accuracy of the predicted pose, two standard metrics are used: Mean Per Joint Position Error (MPJPE) and Mean Per Vertex Error (MPVE), both computed in 3D space after reconstructing the mesh from the predicted pose parameters.
\begin{equation}
    \text{MPJPE} = \frac{1}{B \cdot T \cdot J} \sum_{b,t,j} \left\| \mathbf{p}_{b,t,j}^{\text{(gt)}} - \mathbf{p}_{b,t,j}^{\text{(pred)}} \right\|_2
\end{equation}
where \( \mathbf{p}_{b,t,j} \in \mathbb{R}^3 \) is the 3D position of joint \( j \in \{ \text{neck, head, jaw} \} \).

\begin{equation}
    \text{MPVE} = \frac{1}{B \cdot T \cdot N} \sum_{b,t,n} \left\| \mathbf{v}_{b,t,n}^{\text{(gt)}} - \mathbf{v}_{b,t,n}^{\text{(pred)}} \right\|_2
\end{equation}
where \( \mathbf{v}_{b,t,n} \in \mathbb{R}^3 \) is the 3D position of vertex \( n \) and \( N \) is the number of SMPL mesh vertices (10475).

\subsection{Results}

\begin{table}[h]
\footnotesize
\centering
\caption{Comparison of model variants for head pose tracking using SMPL-X. MPJPE and MPVE Metrics are reported in millimeters and computed against pseudo ground truth from OS-X.}
\begin{tabular}{lcccc|cccc}
\hline
\multirow{2}{*}{\textbf{Model}} & \multicolumn{4}{c}{\textbf{MPJPE (mm)}} & \multicolumn{4}{|c}{\textbf{MPVE (mm)}} \\
\cline{2-9}
 & Neck & Head & Jaw & Avg & Neck & Head & Jaw & Avg \\
\hline
Baseline \cite{ray2024origami} 
& 18.2 & 16.5 & 15.7 & 16.8 & 16.0 & 14.5 & 13.9 & 14.8 \\
Imp2Head 
& 13.0 & 11.3 & 10.8 & 11.7 & 9.0 & 8.2 & 7.9 & 8.4 \\
Imp2Head + $\mathcal{L}_{\text{bio}}$ 
& \textbf{7.5} & \textbf{6.3} & \textbf{6.3} & \textbf{6.7} & \textbf{6.5} & \textbf{5.6} & \textbf{5.6} & \textbf{5.9} \\
\hline
\end{tabular}
\label{tab:mpjpe_results}
\end{table}

\cref{tab:mpjpe_results} summarizes the results of the head pose estimation with different models. 
In this table, the reference ground truth is from the inference of the SOTA vision-based model OS-X, which we denote as the pseudo ground truth.
The baseline model built based on convolutional layers\cite{ray2024origami}, achieves MPJPE and MPVE scores of 16.8 mm and 14.8 mm, respectively. 
The proposed transformer-based model Imp2Head significantly reduces these errors to 11.7 mm (MPJPE) and 8.4 mm (MPVE). 
Further, adding biomechanical constraint terms (Imp2Head + $\mathcal{L}_{\text{bio}}$) leads to the lowest error rates observed, achieving an MPJPE of 6.7 mm and an MPVE of 5.9 mm. 
These results indicate that our proposed Imp2Head model incorporating anatomical biomechanical constraints significantly improves head pose estimation accuracy.

\begin{table}[h]
\footnotesize
\centering
\caption{Comparison of related works and Imp2Head (Proposed approach). Corrected error estimates reflect expected deviation from true 3D ground truth.}
\label{tab:corrected_mpjpe}
\begin{threeparttable}
\begin{tabular}{lcc}
\hline
\textbf{Model} & \textbf{MPJPE (mm)} & \textbf{Notes} \\
\hline

IMUPoser$^1$ \cite{mollyn2023imuposer} &105.0 & 3 IMUs vs MoCap \\
MobilePoser$^1$ \cite{xu2024mobileposer} &126.0 & 3 IMUs vs  MoCap\\
PoseMamba \cite{huang2025posemamba} & 37.5 &Vision-based PD vs MoCap \\
OS-X \cite{lin2023one}  & 24.9 & Vision-based PD vs MoCap \\
\hline
Imp2Head & 6.7  & Imp2Head vs Pseudo GT (Vision-based PD) \\
Imp2Head & \textit{25.9} & Imp2Head vs True GT (MoCap)\\
\hline
\end{tabular}
\begin{tablenotes}
\item[1] The results are from the full body pose estimation. The work \cite{huang2025posemamba} notes a challenge in accurately tracking the head and neck regions, indicating these areas can have lower performance scores compared to limbs in full-body models
\end{tablenotes}
\end{threeparttable}
\end{table}

Since the model is trained and evaluated against a vision-based pose estimation model, which itself is not ground truth but an approximation derived from vision-based inference, the reported MPJPE underestimates the true error. To approximate the model's actual deviation from real 3D ground truth (e.g., MoCap), we apply a Pythagorean error composition, assuming the errors between OS-X and the impedance-based model are independent:

\begin{equation}
    \text{MPJPE}_{\text{Corrected}} \approx \sqrt{(\text{MPJPE}_{\text{OS-X}})^2 + (\text{MPJPE}_{\text{Imp2Head}})^2}
\end{equation}

In \cref{tab:corrected_mpjpe}, we compare our best impedance-based model (Imp2Head + $\mathcal{L}_{\text{bio}}$) against the state-of-the-art vision-based OS-X model \cite{lin2023one} and other related works \cite{mollyn2023imuposer,xu2024mobileposer,huang2025posemamba}. 
Although our model achieves an MPJPE of 6.7 mm against the pseudo-ground truth provided by OS-X, we recognize that OS-X itself has an inherent error of approximately 24.9 mm compared to a true ground truth system like motion capture (MoCap). Using the Pythagorean error composition, which assumes independent errors between the impedance-based method and OS-X, we estimate the corrected MPJPE of our impedance-based model to be approximately 25.9 mm, which is comparable to the
state-of-the-art vision-based system OS-X.

These results affirm that bio-impedance sensing provides a competitive, vision-free, and low-complexity alternative to current vision- and IMU-based head pose estimation methods, suitable for practical wearable scenarios.
\section{Discussion and Limitation}

\begin{figure}[!htbp]
    \centering
\includegraphics[width=0.8\linewidth]{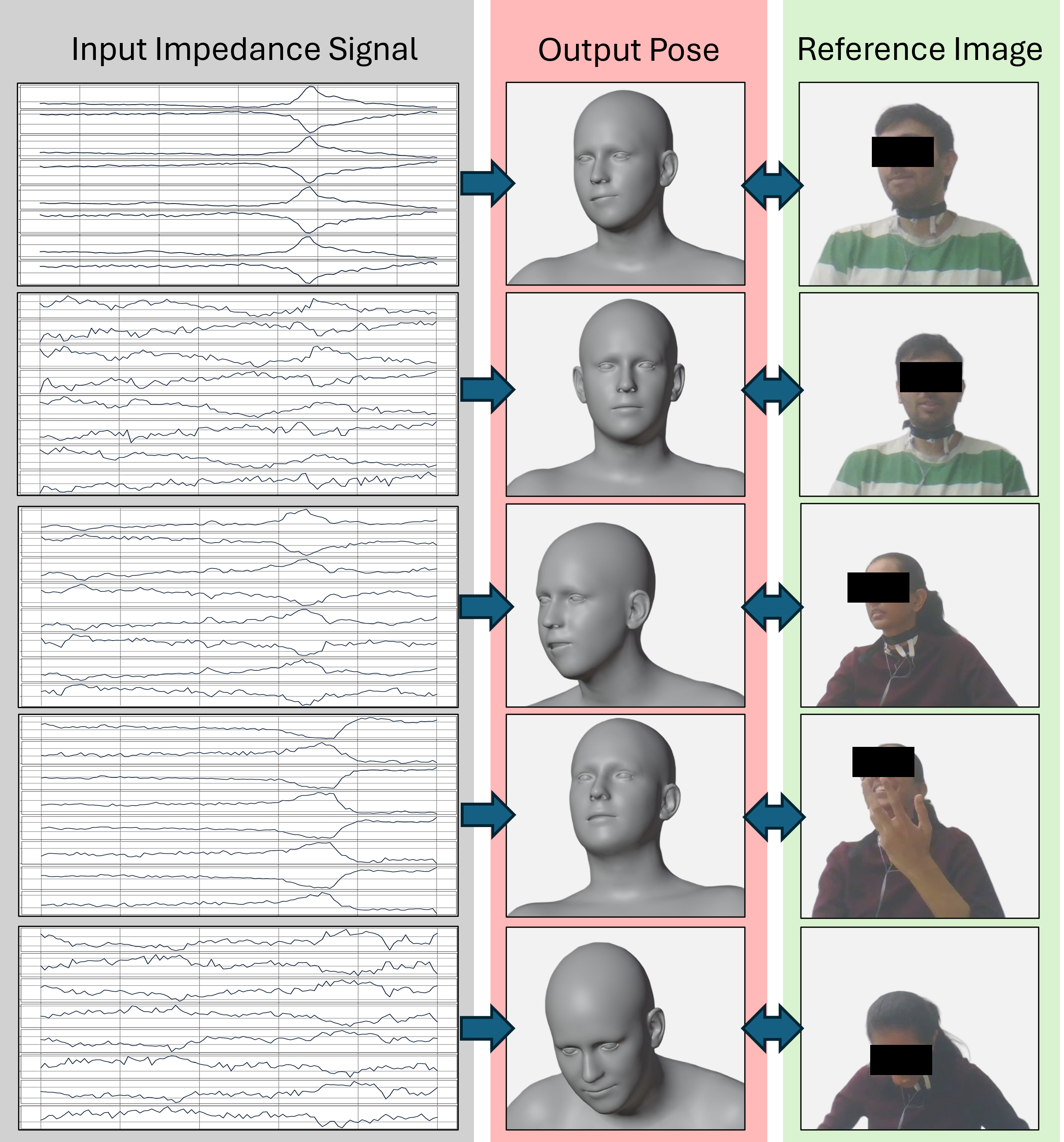}
    \caption{Head pose estimation from neck impedance signals: The left column displays time-series neck bio-impedance signals, the middle column shows the predicted SMPL-X poses derived from these signals using Imp2Head, and the right column provides reference ground-truth images of the subjects in corresponding poses.}
    \label{fig:qualtiative results}
\end{figure}


Our evaluation highlights bio-impedance sensing as a robust, non-intrusive method for wearable head pose estimation. Imp2Head significantly reduces MPJPE and MPVE compared to baselines, benefiting from anatomical constraint integration. Unlike vision-based methods, our approach is resilient to occlusion and lighting changes, making it well-suited for daily use. With a corrected MPJPE of 25.9 mm, comparable to OS-X's 24.9 mm, bio-impedance proves to be a viable alternative. Moreover, it enables jaw pose estimation, a task difficult for IMU, EMG, or capacitive sensors. As shown in \cref{fig:qualtiative results}, our model tracks both head and jaw motion accurately, demonstrating bio-impedance’s unique ability to capture fine anatomical movements beyond the reach of conventional wearables.

Despite promising results, several limitations remain. First, the system relies on a vision-based pose estimator (OS-X) for training and evaluation instead of a gold-standard MoCap system. This introduces potential bias, as errors or drift in the vision pipeline may propagate through training and obscure failure cases. Second, while soft, dry electrodes improve comfort, the system is sensitive to small shifts in placement from neck movement, sweat, or collar adjustments, leading to signal noise and drift. This highlights the need for adaptive preprocessing, auto-calibration, or self-alignment. Third, the model does not account for individual anatomical differences (e.g., neck size, posture, tissue composition), which may affect impedance patterns. Incorporating personalized calibration could improve generalization. Lastly, the small sample size (N=7) limits the robustness and statistical power of the results.

These limitations suggest several directions for future research. Enhancing personalization and adaptability, optimizing the mechanical and material design of the wearable for long-term comfort, and extending the system to support complex or high-dynamic activities represent key opportunities. Additionally, incorporating true MoCap-based ground truth in future evaluations would provide a more rigorous assessment of model fidelity.


\section{Conclusion}


We introduce NeckSense, a bio-impedance-based wearable system for continuous, unobtrusive head pose estimation. It uses multi-channel impedance measurements via soft, dry, reusable electrodes embedded in a lightweight necklace. Our Transformer-based Imp2Head model maps signals to head pose, incorporating biomechanical constraints for enhanced accuracy. Experiments show performance comparable to state-of-the-art vision-based systems, validating bio-impedance as a practical alternative. Future work includes adapting to individual anatomical differences and handling more dynamic motions. Overall, NeckSense advances wearable sensing, enabling reliable and versatile human pose tracking.

\begin{acks}
This work is funded by the European Union under the EIC Pathfinder programme, with the project 101162257 STELEC.
\end{acks}

\bibliographystyle{ACM-Reference-Format}
\bibliography{sample-base}


\end{document}